\documentclass[11pt,a4paper]{article}
\usepackage[hyperref]{naaclhlt2019}
\usepackage{times}
\usepackage{latexsym}
\usepackage{url}

\usepackage{graphicx}
\usepackage{wrapfig}
\usepackage{amsmath}
\usepackage{amssymb}
\usepackage{color,xcolor,colortbl}
\usepackage{enumitem}
\usepackage{algorithm}
\usepackage{algorithmic}
\usepackage[font={small}]{caption}
\usepackage{bm,bbm}
\usepackage{booktabs}
\usepackage{mathtools}
\usepackage{array}
\usepackage{multirow}
\usepackage{soul}
\usepackage{subcaption}
\usepackage{pbox}
\usepackage{pifont}
\usepackage{arydshln}

\newcommand\numberthis{\addtocounter{equation}{1}\tag{\theequation}}

\definecolor{darkblue}{rgb}{0.0, 0.0, 0.55}
\newenvironment{fontppl}{\fontfamily{ppl}\selectfont}{\par} 
\setul{0.5ex}{0.3ex} 
\setulcolor{blue} 
\aclfinalcopy 
\def\aclpaperid{1914} 

\title{Guiding Extractive Summarization with Question-Answering Rewards}

\author{Kristjan Arumae \and Fei Liu\\ 
  Computer Science Department\\
  University of Central Florida, 
  Orlando, FL 32816, USA\\
  {\tt k.arumae@knights.ucf.edu \quad feiliu@cs.ucf.edu}}

\date{}

\begin{document}
\maketitle
\begin{abstract}

Highlighting while reading is a natural behavior for people to track salient content of a document.
It would be desirable to teach an extractive summarizer to do the same.
However, a major obstacle to the development of a supervised summarizer is the lack of ground-truth.
Manual annotation of extraction units is cost-prohibitive, whereas acquiring labels by automatically aligning human abstracts and source documents can yield inferior results.
In this paper we describe a novel framework to guide a supervised, extractive summarization system with question-answering rewards.
We argue that quality summaries should serve as a document surrogate to answer important questions, and such question-answer pairs can be conveniently obtained from human abstracts.
The system learns to promote summaries that are informative, fluent, and perform competitively on question-answering.
Our results compare favorably with those reported by strong summarization baselines as evaluated by automatic metrics and human assessors.

\end{abstract}

\section{Introduction}

Our increasingly digitized lifestyle calls for summarization techniques to produce short and accurate summaries that can be accessed at any time.
These summaries should factually adhere to the content of the source text and present the reader with the key points therein.
Although neural abstractive summarization has shown promising results~\cite{Rush:2015,Nallapati:2016,See:2017}, these methods can have potential drawbacks.
It was revealed that abstracts generated by neural systems sometimes alter or falsify objective details, and introduce new meanings not present in the original text~\cite{Cao:2018}.
Reading these abstracts can lead to misinterpretation of the source materials, which is clearly undesirable.
In this work, we focus on extractive summarization, where the summaries are guaranteed to remain faithful to the original content.
Our system seeks to identify \emph{salient and consecutive sequences of words} from the source document, and highlight them in the text to assist users in browsing and comprehending lengthy documents.
An example is illustrated in Table~\ref{tab:example}.

\begin{table}[t]
\setlength{\tabcolsep}{5pt}
\renewcommand{\arraystretch}{1}
\centering
\begin{scriptsize}
\begin{fontppl}
\begin{tabular}{|l|}
\hline

\begin{minipage}{2in}
\includegraphics[height=0.55in]{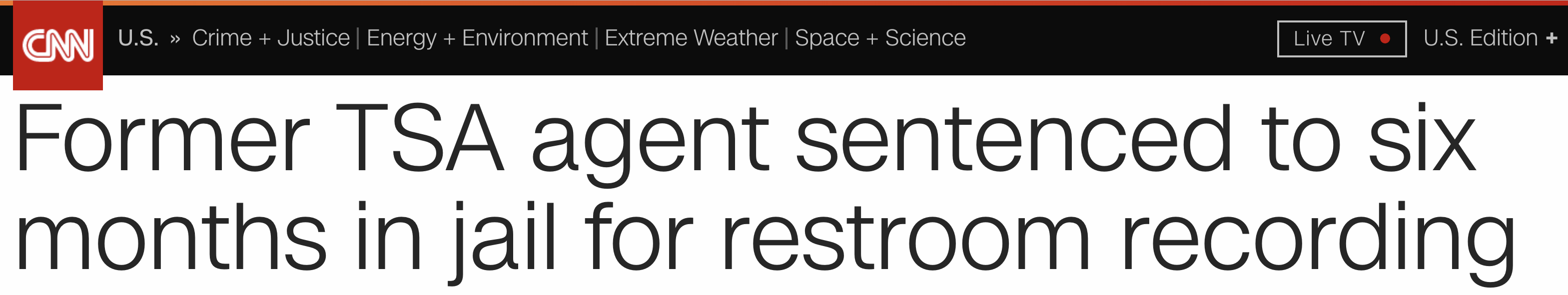}
\end{minipage}
\\
\\
(CNN) \textbf{\textcolor{red}{A judge this week sentenced a \colorbox{green!15}{former TSA agent} to six}} \\ \textbf{\textcolor{red}{months in jail for secretly \colorbox{green!15}{videotaping a female co-worker}}}\\ while she was in the bathroom, prosecutors said.\\[1.2mm]

During the investigation, detectives with \textbf{\textcolor{red}{the Metro Nashville}}\\ 
\textbf{\textcolor{red}{Police Department in Tennessee}} also found that the agent, \\ 33-year-old Daniel Boykin, entered \textbf{\textcolor{red}{the woman's}} home multiple \\ times, where he took videos, photos and other data.\\ [1.2mm]

\colorbox{green!15}{Police found \textbf{\textcolor{red}{more than 90 videos and 1,500 photos of the victim}}}\\ \colorbox{green!15}{on Boykin's phone and computer}.\\[1.2mm]

The victim filed a complaint after seeing images of herself on his\\ phone last year. [...]\\[1.2mm]

\hline
\hline

\textbf{Comprehension Questions (Human Abstract):}\\[1.2mm]
Former \_\_\_\_\_\_\_ Daniel Boykin, 33, videotaped his female co-worker\\ in the restroom, authorities say.\\[1.2mm] 

Authorities say they found 90 videos and 1,500 photos of the victim\\  on \_\_\_\_\_\_\_ and computer.\\[1.2mm]

\hline
\end{tabular}
\end{fontppl}
\end{scriptsize}
\caption{An example extractive summary bolded in the article (top).
Highlighted sections indicate salient segments useful for answering fill-in-the-blank questions generated from human abstracts (bottom).
}
\label{tab:example}
\vspace{-0.2in}
\end{table}

A primary challenge faced by extractive summarizers is the lack of annotated data.
The cost of hiring humans to label a necessary amount of source articles with summary words, good for training a modern classifier, can be prohibitive.
Previous work has exploited using human abstracts to derive labels for extraction units~\cite{Woodsend:2010}.
E.g., a source word is tagged 1 if it appears in the abstract, 0 otherwise.
Although pairs of source articles and human abstracts are abundant, labels derived in this way are not necessarily best since summary saliency can not be easily captured with a rule based categorization.
Considering that human abstracts involve generalization, paraphrasing, and can contain words not present in the source text,
leveraging them to derive labels for extraction units can be suboptimal.
In this work, we investigate a new strategy that seeks to better utilize human abstracts to guide the extraction of summary text units.

We hypothesize that quality extractive summaries should contain informative content so that they can be used as document surrogates to answer important questions, thereby satisfying users' information needs. 
The question-answer pairs can be conveniently developed from human abstracts. 
Our proposed approach identifies answer tokens from each sentence of the human abstract, then replaces each answer token with a blank to create a Cloze-style question-answer pair.
To answer \emph{all} questions ($\approx$human abstract), the system summary must contain content that is semantically close to and collectively resembles the human abstract.

In this paper, we construct an extractive summary by selecting consecutive word sequences from the source document.
To accomplish this we utilize a novel reinforcement learning framework to explore the space of possible extractive summaries and assess each summary using a novel reward function judging the summary's adequacy, fluency, length, and its competency to answer important questions.
The system learns to sample extractive summaries yielding the highest expected rewards, with no pre-derived extraction labels needed. 
This work extends the methodology of Arumae and Liu~\shortcite{Arumae:2018} with new representations of extraction units and thorough experimental evaluation.
The contributions of this research can be summarized as follows:
\begin{itemize}[topsep=3pt,itemsep=-1pt,leftmargin=*]
\item we describe a novel framework generating extractive summaries by selecting consecutive sequences of words from source documents. 
This new system explores various encoding mechanisms, as well as new sampling techniques to capture phrase level data. Such a framework has not been thoroughly investigated in the past;
\item
We conduct a methodical empirical evaluation from the point of view of information saliency.
Rather than solely relying on automatic summarization evaluation methods, we also show the advantages of our system by assessing the summary quality with reading comprehension tasks.
Our summaries compare favorably with the automatic metrics against state of the art, and show promising results against baselines when evaluated by humans for question answering.
\end{itemize}

\section{Related Work}
\label{sec:related}

Extractive summarization has seen growing popularity in the past decades~\cite{Nenkova:2011}.
The methods focus on selecting representative sentences from the document(s) and optionally deleting unimportant sentence constituents to form a summary~\cite{Knight:2002,Radev:2004,Zajic:2007,Martins:2009,Gillick:2009:NAACL,Lin:2010:NAACL,Wang:2013,Li:2013:EMNLP,Li:2014:EMNLP,Hong:2014,Yogatama:2015}.
A majority of the methods are unsupervised.
They estimate sentence importance based on the sentence's length and position in the document, whether the sentence contains topical content and its relationship with other sentences.
The summarization objective is to select a handful of sentences to maximize the coverage of important content while minimizing summary redundancy.
Although unsupervised methods are promising, they cannot benefit from the large-scale training data harvested from the Web~\cite{Sandhaus:2008,Hermann:2015,Grusky:2018}.

Neural extractive summarization has focused primarily on extracting sentences~\cite{Nallapati:2017,Cao:2017,Isonuma:2017,Tarnpradab:2017,Zhou:2018,Kedzie:2018}.
These studies exploit parallel training data consisting of source articles and story highlights (i.e., human abstracts) to create ground-truth labels for sentences.
A neural extractive summarizer learns to predict a binary label for each source sentence indicating if it is to be included in the summary. 
These studies build distributed sentence representations using neural networks~\cite{Cheng:2016,Yasunaga:2017} and use reinforcement learning to optimize the evaluation metric~\cite{Narayan:2018} and improve summary coherence~\cite{Wu:2018}.
However, sentence extraction can be coarse and in many cases, only a part of the sentence is worthy to be added to the summary.
In this study, we perform finer-grained extractive summarization by allowing the system to select consecutive sequences of words rather than sentences to form a summary.

Interestingly, studies reveal that summaries generated by recent neural abstractive systems are, in fact, quite ``extractive.''
Abstractive systems often adopt the encoder-decoder architecture with an attention mechanism~\cite{Rush:2015,Nallapati:2016,Paulus:2017,Guo:2018:ACL,Gehrmann:2018,Lebanoff:2018,Celikyilmaz:2018}.
The encoder condenses a source sequence to a fixed-length vector and the decoder takes the vector as input and generates a summary by predicting one word at a time. 
See, Liu, and Manning~\shortcite{See:2017} suggest that about 35\% of the summary sentences occur in the source documents, and 90\% of summary n-grams appear in the source.
Moreover, the summaries may contain inaccurate factual details and introduce new meanings not present in the original text~\cite{Cao:2018,Song:2018}.
It thus raises concerns as to whether such systems can be used in real-world scenarios to summarize materials such as legal documents.
In this work, we choose to focus on extractive summarization where selected word sequences can be highlighted on the source text to avoid change of meaning.

Our proposed method is inspired by the work of Lei et al.~\shortcite{Lei:2016} who seek to identify rationales from textual input to support sentiment classification and question retrieval. 
Distinct from this previous work, we focus on generating generic document summaries.
We present a novel supervised framework encouraging the selection of consecutive sequences of words to form an extractive summary.
Further, we leverage reinforcement learning to explore the space of possible extractive summaries and promote those that are fluent, adequate, and competent in question answering.
We seek to test the hypothesis that successful summaries can serve as document surrogates to answer important questions, and moreover, ground-truth question-answer pairs can be derived from human abstracts.
In the following section we describe our proposed approach in details.

\section{Our Approach}

Let $\mathcal{S}$ be an extractive summary consisting of text segments selected from a source document $\mathbf{x}$.
The summary can be mapped to a sequence of binary labels $\mathbf{y}$ assigned to document words.
In this section we first present a supervised framework for identifying \emph{consecutive sequences of words} that are summary-worthy, then proceed by describing our question-answering rewards and a deep reinforcement learning framework to guide the selection of summaries so that they can be used as document surrogates to answer important questions.\footnote{We have made our code and models available at \url{https://github.com/ucfnlp/summ_qa_rewards}}

\subsection{Representing an Extraction Unit}
\label{sec:extr_unit}

How best to decompose a source document into a set of text units useful for extractive summarization remains an open problem.
A natural choice is to use words as extraction units.
However, this choice ignores the cohesiveness of text. 
A text chunk (e.g., a prepositional phrase) can be either selected to the summary in its entirety or not at all.
In this paper we experiment with both schemes, using either \emph{words} or \emph{chunks} as extraction units.
When a text chunk is selected in the summary, all its consisting words are selected.
We obtain text chunks by breaking down the sentence constituent parse tree in a top-down manner until each tree fragment governs at most 5 words.
A chunk thus can contain from 1 to 5 words.
Additionally, word level modeling can be considered a special case of chunks where the length of each phrase is 1.
It is important to note that using sentences as extraction units is out of the scope of this paper, because our work focuses on finer-grained extraction units such as words and phrases and this is notably a more challenging task.

The most successful neural models for encoding a piece of text to a fixed-length vector include the recurrent~\cite{Hochreiter:1997} and convolutional neural networks (CNN; Kim et al., 2014\nocite{Kim:2014}), among others.
A recent study by Khandelwal et al.~\shortcite{Khandelwal:2018} reported that the recurrent networks are capable of memorizing a recent context of about 20 tokens and the model is highly sensitive to word order, whereas this is less the case for CNN whose max-pooling operation makes it agnostic to word order.
We implement both networks and are curious to compare their effectiveness at encoding extraction units for summarization. 
\begin{align*}
\{\mathbf{h}_t^e\} &= f_1^{\mbox{\scriptsize Bi-LSTM}}(\mathbf{x})
\numberthis\label{equ:h_t_e_lstm}\\
\mbox{or} \quad \{\mathbf{h}_t^e\} &= f_2^{\mbox{\scriptsize CNN}}(\mathbf{x})
\numberthis\label{equ:h_t_e_cnn}
\end{align*}

Our model first encodes the source document using a bidirectional LSTM with the forward and backward passes (Eq.~(\ref{equ:h_t_e_lstm})).
The representation of the $t$-th source word $\smash{\mathbf{h}_t^e = [\overleftarrow{\mathbf{h}}_t^e || \overrightarrow{\mathbf{h}}_t^e]}$ is the concatenation of the hidden states in both directions.
A chunk is similarly denoted by $\smash{\mathbf{h}_t^e = [\overleftarrow{\mathbf{h}}_t^e || \overrightarrow{\mathbf{h}}_{t+n}^e]}$ where $t$ and $t+n$ are the indices of its beginning and ending words.
In both cases, a fixed-length vector ($\mathbf{h}_t^e \in \mathbb{R}^m$) is created for the word/chunk.
Further, our CNN encoder (Eq.~(\ref{equ:h_t_e_cnn})) uses a sliding window of \{1,3,5,7\} words, corresponding to the kernel sizes, to scan through the source document. 
We apply a number of filters to each window size to extract local features.
The $t$-th source word is represented by the concatenation of feature maps (an $m$-dimensional vector).
To obtain the chunk vector we perform max-pooling over the representations of its consisting words (from $t$ to $t+n$).
In the following we use $\mathbf{h}_t^e$ to denote the vector representation of the $t$-th extraction unit, may it be a word or a chunk, generated using either encoder.

\subsection{Constructing an Extractive Summary}
\label{sec:supervision}

It is desirable to first develop a supervised framework for identifying summary-worthy text segments from a source article.
These segments collectively form an extractive summary to be highlighted on the source text.
The task can be formulated as a sequence labeling problem: a source text unit (a word or chunk) is labelled 1 if it is to be included in the summary and 0 otherwise. 
It is not unusual to develop an auto-regressive model to perform sequence labeling, where the label of the $t$-th extraction unit (${y}_t$) depends on all previous labels ($\mathbf{y}_{<t}$).
Given this hypothesis, we build a framework to extract summary units where the importance of the $t$-th source unit is characterized by its informativeness (encoded in $\mathbf{h}_t^e$), its position in the document, and relationship with the partial summary. 
The details are presented below.

We use a positional embedding ($\mathbf{g}_t$) to signify the position of the $t$-th text unit in the source document.
The position corresponds to the index of the source sentence containing the $t$-th unit, and further, all text units belonging to the same sentence share the same positional embedding.
We apply sinusoidal initialization to the embeddings, following Vaswani et al.~\shortcite{Vaswani:2017}.
Importantly, positional embeddings allow us to inject macro-positional knowledge about words/chunks into a neural summarization framework to offset the natural bias that humans tend to have on putting important content at the beginning of an article.

Next, we build a representation for the partial summary to aid the system in selecting future text units.
The representation $\mathbf{s}_t$ is expected to encode the extraction decisions up to time $t$-1 and it can be realized using a unidirectional LSTM network (Eq.~(\ref{equ:s_t})).
The $t$-th input to the network is represented as $y_{t-1} \otimes \mathbf{h}_{t-1}^e$ where $y_{t-1}$ is a binary label serving as a gating mechanism to control if the semantic content of the previous text unit ($\mathbf{h}_{t-1}^e$) is to be included in the summary (``$\otimes$'' corresponds to elementwise product). 
During training, we apply teacher forcing and $y_{t-1}$ is the ground-truth extraction label for the ($t-1$)-th unit; 
at test time, $y_{t-1}$ is generated on-the-fly by obtaining the label yielding the highest probability according to Eq.~(\ref{equ:y_t}).
In the previous work of Cheng and Lapata~\shortcite{Cheng:2016} and Nallapati et al.~\shortcite{Nallapati:2017}, similar auto-regressive models are developed to identify summary sentences.
Different from the previous work, this study focuses on extracting consecutive sequences of words and chunks from the source document, and the partial summary representation is particularly useful for predicting if the next unit is to be included in the summary to improve summary fluency. 
\begin{align*}
\mathbf{s}_t = f_3^{\mbox{\scriptsize Uni-LSTM}}(\mathbf{s}_{t-1}, y_{t-1} \otimes \mathbf{h}_{t-1}^e)
\label{equ:s_t}\numberthis
\end{align*}

\begin{figure}[t]
\centering
\includegraphics[width=2.7in]{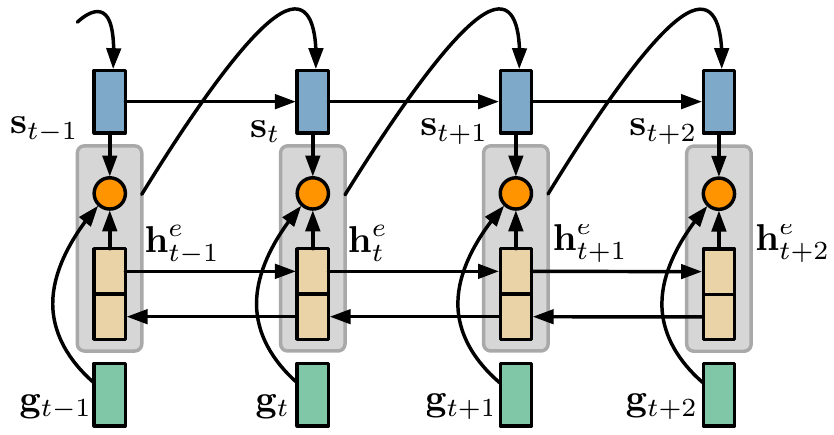}
\caption{
A unidirectional LSTM (blue, Eq.~(\ref{equ:s_t})) encodes the partial summary, while the multilayer perceptron network (orange, Eq.~(\ref{equ:a_t}-\ref{equ:y_t})) utilizes the text unit representation ($\mathbf{h}_t^e$), its positional embedding ($\mathbf{g}_t$), and the partial summary representation ($\mathbf{s}_{t}$) to determine if the $t$-th text unit is to be included in the summary. 
Best viewed in color.
}
\label{fig:architecture}
\vspace{-0.1in}
\end{figure}

Given the partial summary representation ($\mathbf{s}_t$), and representation of the text unit ($\mathbf{h}_t^e$) and its positional encoding ($\mathbf{g}_t$), we employ a multilayer perceptron to predict how likely the unit is to be included in the summary.
This process is described by Eqs.~(\ref{equ:a_t}-\ref{equ:y_t}) and further illustrated in Figure~\ref{fig:architecture}.
\begin{align*}
\mathbf{a}_t &= f^{\mbox{\scriptsize ReLU}}(\mathbf{W}^a [\mathbf{h}_t^e ; \mathbf{g}_t ; \mathbf{s}_t] + \mathbf{b}^a)
\numberthis\label{equ:a_t}\\
p(y_t|\mathbf{y}_{<t}, \mathbf{x}) &= \sigma(\mathbf{w}^y \mathbf{a}_t + b^y)
\numberthis\label{equ:y_t}
\end{align*}

Our model parameters include \{$\mathbf{W}^a$, $\mathbf{b}^a$, $\mathbf{w}^y$, $b^y$\} along with those required by $f_1^{\mbox{\scriptsize Bi-LSTM}}$, $f_2^{\mbox{\scriptsize CNN}}$ and $f_3^{\mbox{\scriptsize Uni-LSTM}}$.
It is possible to train this model in a fully supervised fashion by minimizing the negative log-likelihood of the training data.
We generate ground-truth labels for source text units as follows.
A source \emph{word} receives a label of 1 if both itself and its adjacent word appear in the human abstract (excluding cases where both words are stopwords). 
This heuristic aims to label consecutive source words (2 or more) as summary-worthy, as opposed to picking single words which can be less informative.
A source \emph{text chunk} receives a label of 1 if one of its component words is labelled 1 in the above process.

Because human abstracts are often short and contain novel words not present in source documents, they can be suboptimal for generating ground-truth labels for extraction units.
Only a small portion of the source words (about 8\% in our dataset) are labelled as positive, whereas the vast majority are negative.
Such labels can be ineffective in providing supervision.
In the following section, we investigate a new learning paradigm, which encourages extractive summaries to contain informative content useful for answering important questions, while question-answer pairs can be automatically derived from human abstracts.

\subsection{Using Summaries to Answer Questions}
\label{sec:qa_pairs}

Our hypothesis is that high-quality summaries should contain informative content making them appropriate to serve as document surrogates to satisfy users' information needs. 
We train the extractive summarizer to identify source text units necessary for answering questions, and the question-answer (QA) pairs can be conveniently developed from human abstracts.

To obtain QA pairs, we set an answer token to be either a salient word or a named entity to limit the space of potential answers.
For any sentence in the human abstract, we identify an answer token from it, then replace the answer token with a blank to create a Cloze-style question-answer pair (see Table~\ref{tab:example}).
When a sentence contains multiple answer tokens, a set of QA pairs can be obtained from it.
It is important to note that at least one QA pair should be extracted from each sentence of the abstract. 
Because a system summary is trained to contain content useful for answering \emph{all} questions ($\approx$human abstract), any missing QA pair is likely to cause the summary to be insufficient.

We collect answer tokens using the following methods: (a) we extract a set of entities with tag \{\textsc{per}, \textsc{loc}, \textsc{org}, \textsc{misc}\} from each sentence using the Stanford CoreNLP toolkit~\cite{Manning:2014}; (b) we also identify the \textsc{root} word of each sentence's dependency parse tree along with the sentence's subject/object word, whose type is \{\textsc{nsubj}, \textsc{csubj}, \textsc{obj}, \textsc{iobj}\} (if exists), then add them to the collection of answer tokens.
Further, we prune the answer space by excluding those which appear fewer than 5 times overall.
Having several methods for question construction allows us to explore the answer space properly.
In the results section we perform experiments on root, subject/object, and named entities to see which model provides the best extraction guide.

Given an extractive summary $\mathcal{S}$ containing a set of source text units, and a collection of question-answer pairs $\mathcal{P}=\{(Q_k, e_k^*)\}_{k=1}^K$ related to the source document, we want to develop a mechanism leveraging the extractive summary to answer these questions.
We first encode each question $Q_k$ to a vector representation ($\mathbf{q}_k$). This is achieved by concatenating the last hidden states of the forward/backward passes of a bidirectional LSTM (Eq.~(\ref{equ:q_k})).
Next, we exploit the attention mechanism to locate summary parts that are relevant to answering the $k$-th question.
Given the attention mechanism, an extractive summary $\mathcal{S}$ can be used to answer multiple questions related to the document.
We define $\alpha_{t,k}$ to be the semantic relatedness between the $t$-th source text unit and the $k$-th question.
Following Chen et al.~\shortcite{Chen:2016:CNN}, we introduce a bilinear term to characterize their relationship ($\alpha_{t,k} \propto \mathbf{h}_t^e \mathbf{W}^{\alpha} \mathbf{q}_k$; see Eq.~(\ref{equ:alpha_t_k})). In this process, we consider only those source text units selected in summary $\mathcal{S}$.
Using $\alpha_{t,k}$ as weights, we then compute a context vector $\mathbf{c}_k$ condensing summary content related to the $k$-th question (Eq.~(\ref{equ:c_k})).
\begin{align*}
\mathbf{q}_k &= f_4^{\mbox{\scriptsize Bi-LSTM}}(Q_k)
\numberthis\label{equ:q_k}\\
\alpha_{t,k} &= \frac{\exp(\mathbf{h}_t^e \mathbf{W}^{\alpha} \mathbf{q}_k)}{\sum_{t} \exp(\mathbf{h}_t^e \mathbf{W}^{\alpha} \mathbf{q}_k)}
\numberthis\label{equ:alpha_t_k}\\
\mathbf{c}_k &= \textstyle\sum_{t} \alpha_{t,k} \mathbf{h}_t^e
\numberthis\label{equ:c_k}\\
\mathbf{u}_k & = [\mathbf{c}_k ; \mathbf{q}_k ; |\mathbf{c}_k - \mathbf{q}_k| ; \mathbf{c}_k \otimes \mathbf{q}_k]
\numberthis\label{equ:u_k}
\end{align*}
To predict the most probable answer, we construct a fully-connected network as the output layer.
The input to the network includes a concatenation of the context vector ($\mathbf{c}_k$), question vector ($\mathbf{q}_k$), absolute difference ($|\mathbf{c}_k-\mathbf{q}_k|$) and element-wise product ($\mathbf{c}_k \otimes \mathbf{q}_k$) of the two vectors (Eq.~(\ref{equ:u_k})).
A softmax function is used to estimate a probability distribution over the space of candidate answers: 
\begin{align*}
P(e_k|\mathcal{S}, Q_k) &= \mbox{softmax}(\mathbf{W}^e f^{\mbox{\scriptsize ReLU}}(\mathbf{W}^u \mathbf{u}_k + \mathbf{b}^u)).
\end{align*}

\noindent Such a fully-connected output layer has achieved success on natural language inference~\cite{Mou:2016,Chen:2018:NLI}; here we test its efficacy on answer selection.
The model parameters include $\{\mathbf{W}^{\alpha}, \mathbf{W}^e, \mathbf{W}^u, \mathbf{b}^u\}$ and those of $f_4^{\mbox{\scriptsize Bi-LSTM}}$.

\subsection{A Reinforcement Learning Framework}
\label{sec:reinforce}

In this section we introduce a reinforcement learning framework to explore the space of possible extractive summaries and present a novel reward function to promote summaries that are adequate, fluent, restricted in length, and competent in question answering.
Our reward function consists of four components, whose interpolation weights $\gamma$, $\alpha$, and $\beta$ are tuned on the dev set.
\begin{align*}
\mathcal{R}(\mathbf{y}) = \mathcal{R}_c(\mathbf{y}) + \gamma \mathcal{R}_a(\mathbf{y}) &+ \alpha \mathcal{R}_f(\mathbf{y}) + \beta \mathcal{R}_l(\mathbf{y})
\end{align*}

We define \emph{QA competency} (Eq.~(\ref{equ:r_c})) as the average log-likelihood of correctly answering questions using the system summary ($\mathbf{y}$).
A high-quality system summary is expected to resemble reference summary by using similar wording.
The \emph{adequacy} metric (Eq.~(\ref{equ:r_a})) measures the percentage of overlapping unigrams between the system ($\mathbf{y}$) and reference summary ($\mathbf{y}^*$). 
The \emph{fluency} criterion (Eq.~(\ref{equ:r_f})) encourages consecutive sequences of source words to be selected by preventing many 0/1 switches in the label sequence (i.e., $|y_t - y_{t-1}|$).
Finally, we limit the summary size by setting the ratio of selected words to be close to a threshold $\delta$ (Eq.~(\ref{equ:r_l})).
{\medmuskip=2mu
\thinmuskip=2mu
\thickmuskip=2mu
\nulldelimiterspace=1pt
\scriptspace=1pt
\begin{align*}
\mbox{QA} \quad \mathcal{R}_c(\mathbf{y}) &= \frac{1}{K}\sum_{k=1}^K \log P(e_k^*|\mathbf{y}, Q_k)
\numberthis\label{equ:r_c}\\
\mbox{Adequ.} \quad \mathcal{R}_a(\mathbf{y}) &= \frac{1}{|\mathbf{y}^*|}\mathcal{U}(\mathbf{y}, \mathbf{y}^*)
\numberthis\label{equ:r_a}\\
\mbox{Fluency} \quad \mathcal{R}_f(\mathbf{y}) &= -\sum_{t=2}^{|\mathbf{y}|} |y_t - y_{t-1}|
\numberthis\label{equ:r_f}\\
\mbox{Length} \quad \mathcal{R}_l(\mathbf{y}) &= -\Big|\frac{1}{|\mathbf{y}|}\sum_{t} y_t - \delta\Big|
\numberthis\label{equ:r_l}
\end{align*}}

The reward function $\mathcal{R}(\mathbf{y})$ successfully combines intrinsic measures of summary fluency and adequacy~\cite{Goldstein:2005} with extrinsic measure of summary responsiveness to given questions~\cite{Dang:2006,Murray:2008:MLMI}.
A reinforcement learning agent finds a policy $P(\mathbf{y}|\mathbf{x})$ to maximize the expected reward $\mathbb{E}_{P(\mathbf{y}|\mathbf{x})}[\mathcal{R}(\mathbf{y})]$.
Training the system with policy gradient (Eq.~(\ref{eq:nabla_r})) involves repeatedly sampling an extractive summary $\hat{\mathbf{y}}$ from the source document $\mathbf{x}$.
At time $t$, the agent takes an action by sampling a decision based on $p(y_t|\hat{\mathbf{y}}_{<t}, \mathbf{x})$ (Eq.~(\ref{equ:y_t})) indicating whether the $t$-th source text unit is to be included in the summary.
Once the full summary sequence $\hat{\mathbf{y}}$ is generated, it is compared to the ground-truth sequence to compute the reward $\mathcal{R}(\hat{\mathbf{y}})$.
In this way, reinforcement learning explores the space of extractive summaries and promotes those yielding high rewards.
At inference time, rather than sampling actions from $p(y_t|\mathbf{y}_{<t}, \mathbf{x})$,
we choose $y_t$ that yields the highest probability to generate the system summary $\mathbf{y}$. 
This process is deterministic and no QA is required.
\begin{align*}
& \nabla_\theta \mathbb{E}_{P(\mathbf{y}|\mathbf{x})}[\mathcal{R}(\mathbf{y})]\\
&= \mathbb{E}_{P(\mathbf{y}|\mathbf{x})}[\mathcal{R}(\mathbf{y})\nabla_\theta\log P(\mathbf{y}|\mathbf{x})]\\
&\approx \textstyle\frac{1}{N}\sum_{n=1}^N \mathcal{R}(\hat{\mathbf{y}}^{(n)}) \nabla_\theta \log P(\hat{\mathbf{y}}^{(n)}|\mathbf{x})
\numberthis\label{eq:nabla_r}
\end{align*}

\section{Experiments}
\label{sec:experiments}

We proceed by discussing the dataset and settings, comparison systems, and experimental results obtained through both automatic metrics and human evaluation in a reading comprehension setting.

\subsection{Dataset and Settings}
\label{sec:data}

Our goal is to build an extractive summarizer identifying important textual segments from source articles.
To investigate the effectiveness of the proposed approach, we conduct experiments on the CNN/Daily Mail dataset using a version provided by See et al.~\shortcite{See:2017}. 
The reference summaries of this dataset were created by human editors exhibiting a moderate degree of extractiveness.
E.g., 83\% of summary unigrams and 45\% of bigrams appear in source articles~\cite{Narayan:2018:EMNLP}.
On average, a CNN article contains 761 words / 34 sentences and a DM article contains 653 words / 29 sentences.
We report results respectively for the CNN and DM portion of the dataset.

Our hyperparameter settings are as follows.
We set the hidden state dimension of the LSTM to be 256 in either direction. 
A bidirectional LSTM $f_1^{\mbox{\scriptsize Bi-LSTM}}(\cdot)$ produces a 512-dimensional vector for each content word.
Similarly, $f_4^{\mbox{\scriptsize Bi-LSTM}}(\cdot)$ generates a question vector $\mathbf{q}_k$ of the same size.
Our CNN encoder $f_2^{\mbox{\scriptsize CNN}}(\cdot)$ uses multiple window sizes of  $\{1,3,5,7\}$ and 128 filters per window size.
$\mathbf{h}_t^e$ is thus a 512-dimensional vector using either CNN or LSTM encoder.
We set the hidden state dimension of $\mathbf{s}_t$ to be 128. 
We also use 100-dimensional word embeddings~\cite{pennington2014glove} and sinusoidal positional encodings~\cite{Vaswani:2017} of 30 dimensions. 

The maximum article length is set to 400 words. 
Compared to the study of Arumae and Liu~\shortcite{Arumae:2018}, we expand the search space dramatically from 100 to 400 words, which poses a challenge to the RL-based summarizers.
We associate each article with at most 10 QA pairs ($K$=10) and use them to guide the extraction of summary segments.
We apply mini-batch training with Adam optimizer~\cite{kingma2014adam}, where a mini-batch contains 128 articles and their QA pairs.
The summary ratio $\delta$ is set to 0.15, yielding extractive summaries of about 60 words.
Following Arumae and Liu~\shortcite{Arumae:2018}, we set hyperparameters $\beta = 2\alpha$; $\alpha$ and $\gamma$ are tuned on the dev set using grid search.

\begin{table}
\setlength{\tabcolsep}{6.3pt}
\renewcommand{\arraystretch}{1.1}
\centering
\begin{small}
\begin{tabular}{lrrrr}
& & \multicolumn{3}{c}{\textbf{CNN}} \\
\textbf{System} & \textbf{\#Ans.} & \textbf{R-1} & \textbf{R-2} & \textbf{R-L}\\
\toprule
Lead-3 & -- & 28.8 & 11.5 & 19.3 \\
PointerGen+Cov. & -- & 29.9 & 10.9 & 21.1\\
Graph Attn. & -- & 30.3 & 9.8 & 20.0\\
LexRank & -- & 26.1 & 9.6 & 17.7 \\
SumBasic & -- & 22.9 & 5.5 & 14.8 \\
KLSum & -- & 20.7 & 5.9 & 13.7 \\
Distraction-M3 & -- & 27.1 & 8.2 & 18.7 \\
\midrule
QASumm+NoQ  & 0 & 16.38 & 7.25 & 11.30 \\
QASumm+SUBJ/OBJ & 9,893 & 26.16 & 8.97 & 18.24\\
QASumm+ROOT  & 3,678 & 26.67 & 9.19 & 18.76 \\
QASumm+NER & 6,167 & \textbf{27.38} & \textbf{9.38} & \textbf{19.02} \\
\bottomrule
\end{tabular}
\end{small}
\caption{
Summarization results on CNN test set. 
Summaries are evaluated at their full-length by ROUGE F$_1$-scores.}
\label{tab:result_summ_cnn}
\end{table}

\begin{table}
\setlength{\tabcolsep}{5.5pt}
\renewcommand{\arraystretch}{1.1}
\centering
\begin{small}
\begin{tabular}{lrrrr}
& & \multicolumn{3}{c}{\textbf{Daily Mail}}\\
\textbf{System} & \textbf{\#Ans.} & \textbf{R-1} & \textbf{R-2} & \textbf{R-L}\\
\toprule
Lead-3 & -- & 22.5 & 9.3 & 20.0\\
PointerGen+Cov. & -- & 31.2 & 17.0 & 28.9\\
Graph Attn. & -- & 27.4 &  11.3 & 15.1\\
NN-WE & -- & 15.7 & 6.4 &  9.8 \\
NN-SE & -- & 22.7 & 8.5 & 12.5 \\
SummaRuNNer & -- & 26.2 & 10.8 & 14.4 \\
\midrule
QASumm+NoQ & 0 & 22.26 & 9.16 & 19.78\\
QASumm+SUBJ/OBJ  & 19,151 & 23.38 & 9.54 & 20.14\\
QASumm+ROOT & 5,498 & \textbf{26.87} & \textbf{11.97} & \textbf{23.07}\\
QASumm+NER & 15,342 & 25.74 & 11.89 & 22.38\\
\bottomrule
\end{tabular}
\end{small}
\caption{
Summarization results on DM test set. 
To ensure a fair comparison, we follow the convention to report ROUGE recall scores evaluated at 75 bytes.}
\label{tab:result_summ_dm}
\vspace{-0.1in}
\end{table}

\subsection{Experimental Results}
\label{sec:results}

\vspace{0.05in}
\noindent \textbf{Comparison systems}\quad
We compare our method with a number of extractive and abstractive systems that have reported results on the CNN/DM datasets.
We consider non-neural approaches that extract sentences from the source article to form a summary. 
These include \emph{LexRank}~\cite{Radev:2004}, \emph{SumBasic}~\cite{Vanderwende:2007}, and \emph{KLSum}~\cite{Haghighi:2009}.
Such methods treat sentences as bags of words, and then select sentences containing topically important words.  
We further include the Lead-3 baseline that extracts the first 3 sentences from any given article.
The method has been shown to be a strong baseline for summarizing news articles.

Neural extractive approaches focus on learning vector representations for sentences and words, then performing extraction based on the learned representations. 
Cheng et al.~\shortcite{Cheng:2016} describe a neural network method composed of a hierarchical document encoder and an attention-based extractor. 
The system has two variants: \emph{NN-WE} extracts words from the source article and \emph{NN-SE} extracts sentences.
\emph{SummaRuNNer}~\cite{Nallapati:2017} presents an autoregressive sequence labeling method based on recurrent neural networks.
It selects summary sentences based on their content, salience, position, and novelty representations. 

Abstractive summarization methods are not directly comparable to our approach, but we choose to include three systems that report results respectively for CNN and DM datasets.
\emph{Distraction-M3}~\cite{Chen:2016} trains the summarization system to distract its attention to traverse different regions of the source article.
\emph{Graph attention}~\cite{Tan:2017} introduces a graph-based attention mechanism to enhance the encoder-decoder framework.
\emph{PointerGen+Cov.}~\cite{See:2017} allows the system to not only copy words from the source text but also generate summary words by selecting them from a vocabulary.
Abstractive methods can thus introduce new words to the summary that are not present in the source article. 
However, system summaries may change the meaning of the original texts due to this flexibility.

\begin{table*}
\setlength{\tabcolsep}{7pt}
\renewcommand{\arraystretch}{1.1}
\centering
\begin{small}
\begin{tabular}{lccc|ccc|ccc|ccc}
\toprule
& \multicolumn{3}{c}{\textbf{NoText}} & \multicolumn{3}{c}{\textbf{QASumm+NoQ}} & \multicolumn{3}{c}{\textbf{GoldSumm}} & \multicolumn{3}{c}{\textbf{FullText}} \\
& \textbf{Train} & \textbf{Dev} & \textbf{Gap} & \textbf{Train} & \textbf{Dev} & \textbf{Gap} & \textbf{Train} & \textbf{Dev} & \textbf{Gap} & \textbf{Train} & \textbf{Dev} & \textbf{Gap}\\
\midrule
SUBJ/OBJ & 49.7 & 24.4 & 25.3 & 55.9 & 31.2 & 24.7 & 69.3 & 48.6 & 20.7 & 67.6 & 43.3 & 24.3\\
ROOT & {\cellcolor[gray]{.8}} 68.1 & {\cellcolor[gray]{.8}} 34.9 & 33.2 & {\cellcolor[gray]{.8}} 71.6 & {\cellcolor[gray]{.8}} 36.3 & 35.3 & 76.9 & 44.9 & 32.0 & 76.0 & 35.7 & 40.3 \\
NER & 61.0 & 15.8 & 45.2 & 66.0 & 32.7 & 33.3 & {\cellcolor[gray]{.8}} 85.2 & {\cellcolor[gray]{.8}} 54.0 & 31.2 & {\cellcolor[gray]{.8}} 82.4 & {\cellcolor[gray]{.8}} 46.3 & 36.1\\
\bottomrule
\end{tabular}
\end{small}
\caption{Question-answering accuracies using different types of QA pairs (ROOT, SUBJ/OBJ, NER) and different source input (NoText, QASumm+NoQ, GoldSumm, and FullText) as the basis for predicting answers.}
\label{tab:results_qa}
\vspace{-0.1in}
\end{table*}

\vspace{0.05in}
\noindent \textbf{Results}\quad
We present summarization results of various systems in Tables~\ref{tab:result_summ_cnn} and \ref{tab:result_summ_dm}, evaluated on the standard CNN/DM test sets by R-1, R-2, and R-L metrics~\cite{Lin:2004}, which respectively measure the overlap of unigrams, bigrams, and longest common subsequences between system and reference summaries.
We investigate four variants of our method:
\emph{QASumm+NoQ} does not utilize any question-answer pairs during training.
It extracts summary text chunks by learning from ground-truth labels (\S\ref{sec:supervision}) and the chunks are encoded by $f_1^{\mbox{\scriptsize Bi-LSTM}}$.
Other variants initialize their models using pretrained parameters from QASumm+NoQ, then integrate the reinforcement learning objective (\S\ref{sec:reinforce}) to exploit the space of possible extractive summaries and reward those that are useful for answering questions.
We consider three types of QA pairs: the answer token is the root of a sentence dependency parse tree (+ROOT), a subject or object (+SUBJ/OBJ), or an entity found in the sentence (+NER).
In all cases, the question is generated by replacing the answer token with a blank symbol.

As illustrated in Tables~\ref{tab:result_summ_cnn} and \ref{tab:result_summ_dm}, our QASumm methods with reinforcement learning (+ROOT, +SUBJ/OBJ, +NER) perform competitively with strong baselines.
They outperform the counterpart \emph{QASumm+NoQ} that makes no use of the QA pairs by a substantial margin.
They outperform or perform at a comparable level to state-of-the-art published systems on the CNN/DM datasets but are generally inferior to PointerGen.
We observe that exacting summary chunks is highly desirable in real-world applications as it provides a mechanism to generate concise summaries.
Nonetheless, accurately identifying summary chunks is challenging because the search space is vast and spuriousness arises in chunking sentences.
Cheng and Lapata~\shortcite{Cheng:2016} report a substantial performance drop when adapting their system to extract words.
Our QASumm methods focusing on chunk extraction perform on par with competitive systems that extract whole sentences.
We additionally present human evaluation results of summary usefulness for a reading comprehension task in \S\ref{sec:human_eval}.

In Tables~\ref{tab:result_summ_cnn} and \ref{tab:result_summ_dm}, we further show the number of unique answers per QA type.
We find that the ROOT-type QA pairs have the least number of unique answers. 
They are often main verbs of sentences.
In contrast, the SUBJ/OBJ-type has the most number of answers.
They are subjects and objects of sentences and correspond to an open class of content words. 
The NER-type has a moderate number of answers compared to others.
Note that all answer tokens have been filtered by frequency; those appearing less than 5 times in the dataset are removed to avoid overfitting. 

Among variants of the QASumm method, 
we find that \emph{QASumm+ROOT} achieves the highest scores on DM dataset. 
\emph{QASumm+NER} performs consistently well on both CNN and DM datasets, suggesting QA pairs of this type are effective in guiding the system to extract summary chunks.
We conjecture that maintaining a moderate number of answers is important to maximize performance.
To answer questions with missing entities, the summary is encouraged to contain similar content as the question body.
Because questions are derived from the human abstract, this in turn requires the system summary to carry similar semantic content as the human abstract.

\vspace{0.05in}
\noindent \textbf{Question-answering accuracy}\quad
We next dive into the QA component of our system to investigate question-answering performance when different types of summaries and QA pairs are supplied to the system (\S\ref{sec:qa_pairs}).
Given a question, the system predicts an answer using an extractive summary as the source input.
Intuitively, an informative summary can lead to high QA accuracy, as the summary content serves well as the basis for predicting answers.
With the same summary as input, certain types of questions can be more difficult to answer than others, and the system must rely heavily on the summary to gauge correct answers.

We compare various types of summaries.
These include
(a) \emph{QASumm+NoQ} which extracts summary chunks without requiring QA pairs;
and (b) \emph{GoldSumm}, which are gold-standard extractive summaries generated by collecting source words appearing in human summaries.
We further consider \emph{NoText} and \emph{FullText}, corresponding to using no source text or the full source article as input.
They represent the two extremes.
In all cases the QA component (\S\ref{sec:qa_pairs}) is trained on the training set and we report QA accuracies on the dev set.

In Table~\ref{tab:results_qa},
we observe that question-answering with \emph{GoldSumm} performs the best for all QA types.
It outperforms the scenarios using \emph{FullText} as the source input. 
This indicates that distilled information contained in a high-quality summary can be useful for answering questions, as searching for answers in a succinct summary can be more efficient than that in a full article.
Moreover, we observe that the performance of \emph{QASumm+NoQ} is in between \emph{NoText} and \emph{GoldSumm} for all answer types. 
The results suggest that extractive summaries with even modest ROUGE scores can prove useful for question-answering.
Regarding different types of QA pairs,
we find that the ROOT-type can achieve high QA accuracy when using \emph{NoText} input.
It suggests that ROOT answers can to some extent be predicted based on the question context.
The NER-type QA pairs work the best for both \emph{GoldSumm} and \emph{FullText}, likely because the source texts contain necessary entities required to correctly answer those questions. 
We also find the SUBJ/OBJ-type QA pairs have the smallest gap between train/dev accuracies, despite that they have a large answer space.
Based on the analysis we would suggest future work to consider using NER-based QA pairs as they encourage the summaries to contain salient source content and be informative.

\begin{figure}[t]
\centering
\includegraphics[width=2.3in]{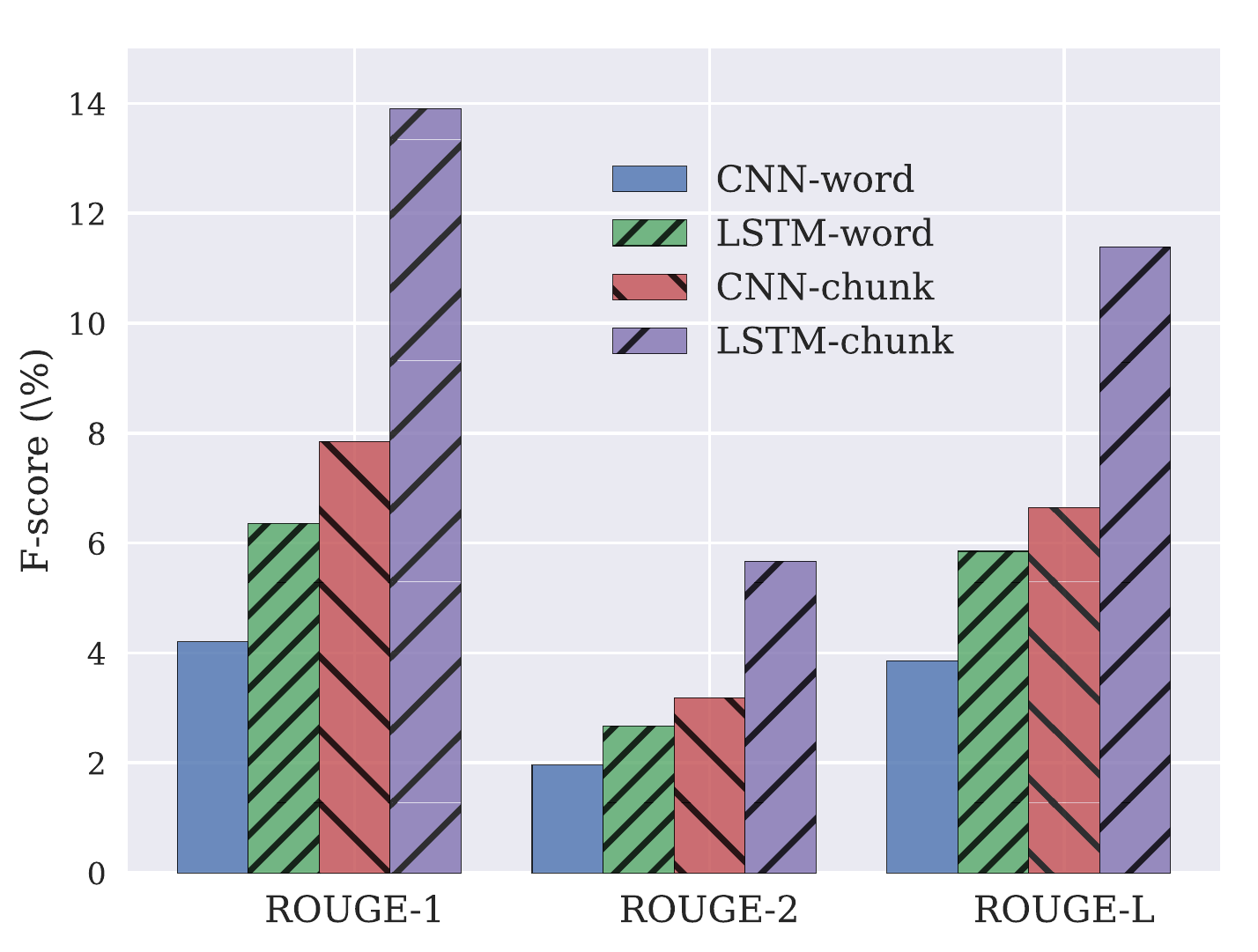}
\caption{Summarization results using $f_1^{\mbox{\scriptsize \textsf{LSTM}}}$ or $f_2^{\mbox{\scriptsize \textsf{CNN}}}$ encoder with word/chunk as the extraction unit.}
\label{fig:extr_unit}
\vspace{-0.15in}
\end{figure}

\vspace{0.05in}
\noindent\textbf{Extraction units}\quad
We finally compare the performance of using either words or chunks as extraction units (\S\ref{sec:extr_unit}).
The chunks are obtained by breaking down sentence constituent parse trees in a top-down manner until all tree fragments contain 5 words or less.
We observe that 70\% of the chunks are 1-grams, and 2/3/4/5-grams are 9\%, 7\%, 6\%, and 8\% respectively.
We compare the bidirectional LSTM ($f_1^{\mbox{\scriptsize \textsf{LSTM}}}$) and CNN ($f_2^{\mbox{\scriptsize \textsf{CNN}}}$) encoders for their effectiveness on generating representations for extraction units.
Figure~\ref{fig:extr_unit} presents the results of the QASumm+NoQ system under various settings.
We find that extracting chunks performs superior, and combining chunks with LSTM representations yield the highest scores.

\subsection{Human Evaluation}
\label{sec:human_eval}

Testing the usefulness of an extractive system driven by reading comprehension is not inherently measured by automatic metrics (i.e. ROUGE).
We conducted a human evaluation to assess whether the highlighted summaries contribute to document understanding.
Similar to our training paradigm we presented each participant with the document and three fill-in-the-blank questions created from the human abstracts.
It was guaranteed that each question was from a unique human abstract to avoid seeing the answer adjacent to the same template.
The missing section was randomly generated to be either the root word, the subject or object of the sentence, or a named entity.
We compare our reinforced extracted summary (presented as a bold overlay to the document), against our supervised method (section 3.2), abstractive summaries generated by See et al. \shortcite{See:2017}, and the human abstracts in full.
Additionally we asked the participants to rate the quality of the summary presented (1-5, with 5 being most informative).
We utilized Amazon Mechanical Turk, and conducted an experiment where we sampled 80 documents from the CNN test set.
The articles were evenly split across the four competing systems, and each HIT was completed by 5 turkers.
Upon completion the data was analyzed manually for accuracy since turkers entered each answer as free text, and to remove any meaningless datapoints.

\begin{table}[t]
\setlength{\tabcolsep}{7pt}
\renewcommand{\arraystretch}{1.2}
\centering
\begin{small}
\begin{tabular}{lrrr}
\toprule
\textbf{Summary} & \textbf{Time} &  \textbf{Accuracy} & \textbf{Inform.} \\
\midrule
Human & 69.5s & 87.3 & 4.23\\
QASumm+NoQ & 87.9s & 53.7 & 1.76\\
Pointer+Cov. & 100.9s & 52.3 & 2.14\\
QASumm+NER & 95.0s & \textbf{62.3} & \textbf{2.14}\\
\bottomrule
\end{tabular}
\end{small}
\caption{Amazon mechanical turk experiments. Human abstracts were the goldstandard summaries, Pointer+Cov. were summaries generated by See et al. \shortcite{See:2017}.  Our systems tested were the supervised extractor, and our full model (NER).}
\label{tab:amt}
\vspace{-0.15in}
\end{table}

Table \ref{tab:amt} shows the average time (in seconds) to complete a single question, the overall accuracy of the participants, and the informativeness of a given summary type.
Excluding the use of human abstracts, all systems resulted in similar performance times.
However we observe a large margin in QA accuracy in our full system compared to the abstractive and our supervised approach. 
Although participants rated the informativeness of the summaries to be the same our systems yielded a higher performance.
This strongly indicates that having a system which makes using of document comprehension has a tangible effect when applied towards a real-world task.

\section{Conclusion}

We exploited an extractive summarization framework using deep reinforcement learning to identify consecutive word sequences from a document to form an extractive summary.
Our reward function promotes adequate and fluent summaries that can serve as document surrogates to answer important questions, directly addressing users' information needs.
Experimental results on benchmark datasets demonstrated the efficacy of our proposed method over state-of-the-art baselines, assessed by both automatic metrics and human evaluators.



\bibliography{summ,abs_summ,fei}
\bibliographystyle{acl_natbib}



\end{document}